\documentclass[10pt,twocolumn,letterpaper]{article}
\usepackage{booktabs}
\usepackage{makecell}
\usepackage{wacv}              % To produce the CAMERA-READY version
\definecolor{wacvblue}{rgb}{0.21,0.49,0.74}
\usepackage[pagebackref,breaklinks,colorlinks,allcolors=wacvblue]{hyperref}
\usepackage{booktabs}

\def\wacvPaperID{} % *** Enter the WACV Paper ID here
\def\confName{WACV}
\def\confYear{2026}

\usepackage{algorithm}
\usepackage{algorithmic}

\usepackage[table]{xcolor}
\usepackage{adjustbox}

\title{
Spatial-Conditioned Reasoning in Long-Egocentric Videos
}

\author{James Tribble$^1$, Hao Wang$^1$, Si-En Hong$^1$, Chaoyi Zhou$^1$, Ashish Bastola$^1$,\\ Siyu Huang$^1$, and Abolfazl Razi$^1$\thanks{Research reported in this publication was supported in part by the NSF and SC EPSCoR Program under award number (NSF Award \# OIA-2242812 and SC EPSCoR 26-CRP03). The views, perspective, and content do not necessarily represent the official views of the SC EPSCoR Program nor those of the NSF.}\\
$^1$Clemson University, SC, USA\\
{\tt\small \{jjtribb, hao9, sienh, chaoyiz, abastol, siyuh, arazi\}@clemson.edu}}

\begin{document}
\maketitle

\begin{abstract}
% Visual navigation, traditionally developed for autonomous robotics, offers a significant advantage for assistive technologies for people with visual impairments (PVI). Existing Vision Language Models (VLMs) excel at open vocabulary understanding and Chain-of-Thought (CoT) reasoning for social compliance; their application for real-time, safe outdoor navigation assistance for PVI remains under-explored. Our work introduces a benchmark based on the Google SANPO dataset to evaluate the potential for various VLMs in assisting PVI with navigation. We evaluate VLM3R, a spatial VLM, against traditional VLM models. Our preliminary results demonstrate that while untrained VLM3R slightly underperforms in raw accuracy on novel tasks, it has superior long-term potential for identifying obstruction-less paths and maintaining social awareness. Furthermore, we demonstrate that low-parameter models can identify environmental patterns more effectively than larger models, albeit at the expense of lower overall accuracy.

Long-horizon egocentric video presents significant challenges for visual navigation due to viewpoint drift and the absence of persistent geometric context. Although recent vision–language models (VLMs) perform well on image and short-video reasoning, their spatial reasoning capability in long egocentric sequences remains limited. In this work, we study how explicit spatial signals influence VLM-based video understanding without modifying model architectures or inference procedures. We introduce Sanpo-D, a fine-grained re-annotation of the Google Sanpo dataset, and benchmark multiple VLMs on navigation-oriented spatial queries. To examine input-level inductive bias, we further fuse depth maps with RGB frames and evaluate their impact on spatial reasoning. Our results reveal a trade-off between general-purpose accuracy and spatial specialization, showing that depth-aware and spatially grounded representations can improve performance on safety-critical tasks such as pedestrian and obstruction detection. 
Our dataset and annotation are available at: 
\url{https://huggingface.co/datasets/Kevius/sanpo_annotations}

\vspace{-0.25in}
\end{abstract}
%-------------------------------------------------------------------------
\section{Introduction}
\label{sec:intro}

% Long-horizon egocentric video presents a challenging setting for visual reasoning due to the absence of loop closure and the lack of persistent geometric context. In this work, we investigate how explicit geometric representations extracted from long egocentric videos can influence vision-language model (VLM) reasoning without altering existing inference frameworks or model parameters.

\begin{figure}[htbp]
    \centering
    \centerline{\includegraphics[width=1\linewidth]{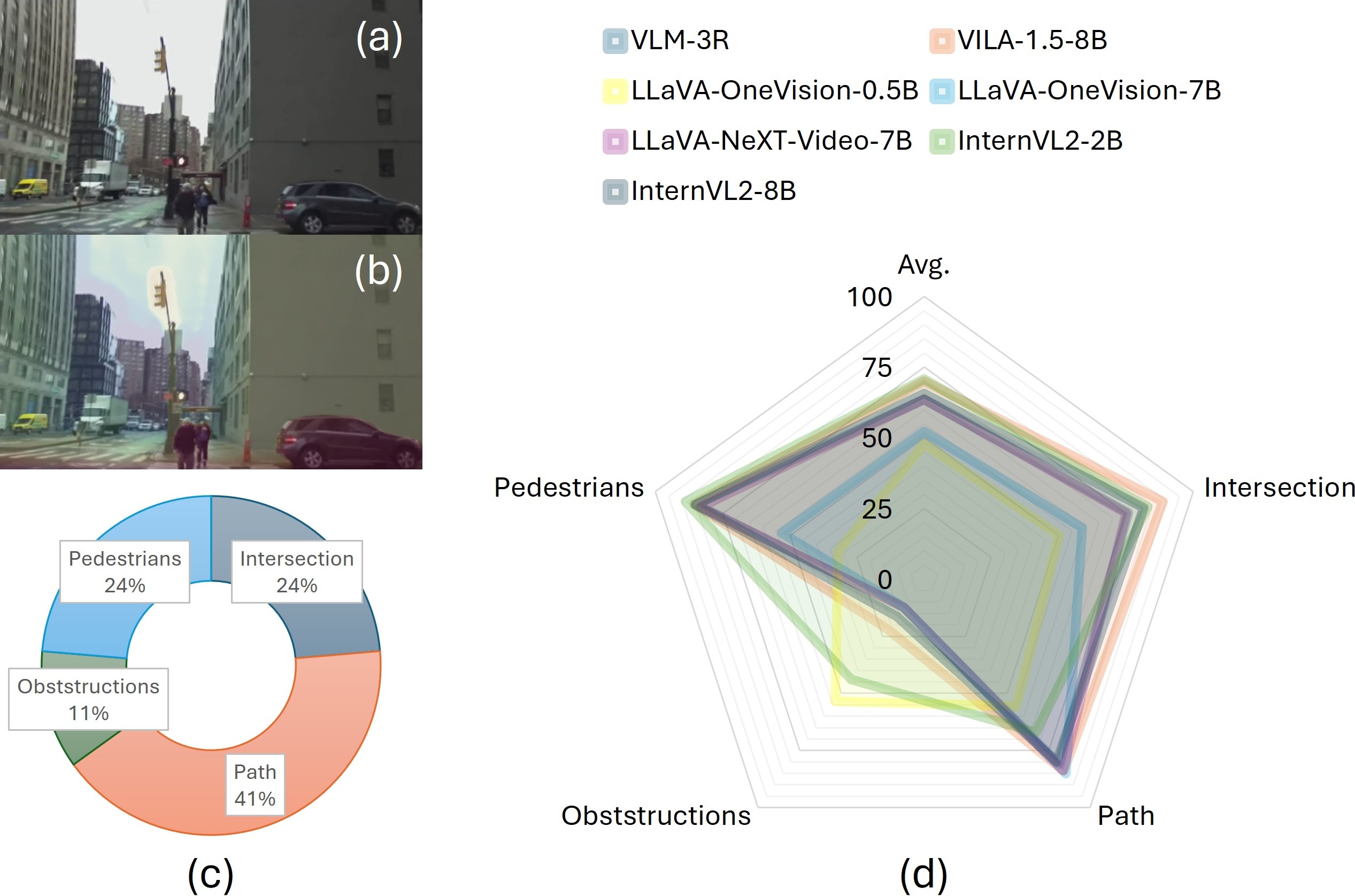}}
    \caption{VLM benchmark on the proposed Sanpo-D dataset.
(a) A raw RGB frame from the Sanpo dataset~\cite{SANPO}.
(b) The corresponding RGB–depth fused frame.
(c) The distribution of annotation types in the Sanpo-D benchmark.
(d) Benchmark results of different VLMs evaluated on the proposed Sanpo-D dataset.}
    \label{fig:cover}
    \vspace{-0.2in}
\end{figure}

\begin{figure*}[htbp]
    \centering
    \centerline{\includegraphics[width=1\linewidth]{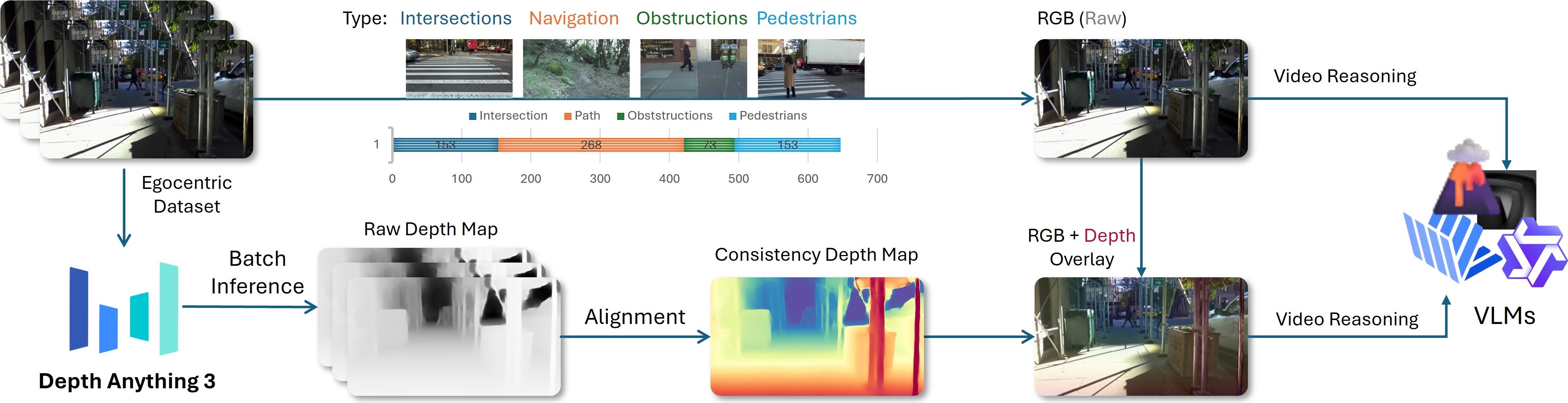}}
    \caption{Overall framework of data annotation and VLM prompting}
    \label{fig:frame}
\end{figure*}

%I can expand the intro further, especially with the current solutions, but all the key points are covered here. 
%introduce the problem/ visual navigation background

% Vision is a key factor for humans to navigate safely in outdoor environments. When a person jogs repeatedly in a new environment, they gradually develop a mental map of safety hazards, landmarks, and obstructions. PVI lacks the ability to perceive their environment visually, which hinders their ability to navigate safely. We are concerned with the reliability and accuracy of VLMs for assisting a blind user with navigation.

Visual navigation constitutes a class of spatial reasoning that humans acquire implicitly through perception and experience, yet it remains a fundamental challenge for modern AI systems, including robotics and vision–language models. 
Although humans routinely perceive spatial layout, proximity, and navigational affordances without explicit modeling, reproducing this capability in computational systems requires expressive representations and reasoning mechanisms.

% Spatial visual understanding is fundamental to safe and effective navigation in outdoor environments. While visual navigation not only recognizes objects, it also reasons their spatial relationships, proximity, and navigational relevance within a scene, as humans generally rely on visual features to reason about spatial layout, obstacles, landmarks, and relative distances, gradually forming an internal representation of their surroundings through repeated exposure. 

Unlike short clips or third-person videos, long-horizon egocentric video presents a particularly challenging setting for spatial visual reasoning, as egocentric recordings span extended durations with unstructured motion, frequent viewpoint drift, and no loop closure, resulting in the absence of persistent geometric context over time. Spatial layouts continuously evolve, observations are partial and transient, and critical navigational cues may appear only briefly. These characteristics make it difficult to maintain a coherent spatial understanding when relying solely on image-based observations.

Many existing visual navigation systems lack the reasoning capability to support query-driven understanding of the surrounding environment and to provide generalized answers beyond predefined instructions. Traditional navigation solutions primarily rely on GPS signals, audio, or geometric routing strategies to deliver turn-based guidance or point-of-interest information. While effective for basic navigation, these systems typically operate on simplified spatial representations and do not account for the type of obstruction, safety hazards, or broader contextual information along a route. 
% In long-horizon egocentric video settings, where global maps are unavailable and spatial context must be inferred incrementally over time, such limitations become more pronounced and restrict the applicability of existing approaches to complex, open-ended environments.

Vision-language models (VLMs), such as VILA, LLaVA-OneVision, LLaVA-NeXT-Video, and InternVL2 \cite{liu2024llavanext, lin2024vilapretrainingvisuallanguage, li2024llavaonevisioneasyvisualtask}, are designed to process video inputs and perform general video understanding, but they primarily rely on implicit appearance-based features and often struggle with tasks that require explicit and persistent spatial reasoning over time. 
Spatially grounded VLMs, such as VLM3R \cite{fan2025vlm3rvisionlanguagemodelsaugmented}, aim to address this limitation by explicitly constructing and tokenizing point-based representations of the scene, thereby incorporating geometric structure into the model input and improving spatio-temporal reasoning capabilities. 
However, in the setting of long-egocentric video, where loop closure is often absent and spatial context must be inferred incrementally from fragmented observations, it is challenging to reason over extended egocentric sequences, as shown in Figure \ref{fig:cover}.

To better understand this issue, we perform a fine-grained annotation of the Google Sanpo dataset \cite{SANPO} (647 samples) and conduct a detailed evaluation across different VLMs. In particular, we hypothesize that the input representation of VLMs introduces a strong inductive bias for video understanding. 
To examine this effect, we employ a metric-level depth estimation model to augment the RGB input with depth maps by fusing depth information into the visual input \cite{DA3}, and evaluate whether this modification influences the spatial reasoning behavior of VLMs without altering their underlying architectures or inference procedures, as illustrated in Figure~\ref{fig:frame}.

We make the following contributions:
\begin{itemize}
    \item We provide a fine-grained annotation of the Google Sanpo dataset \cite{SANPO}, focusing on spatially grounded queries relevant to visual navigation in long-horizon egocentric video.
    \item We conduct a detailed evaluation of multiple vision-language models on visual navigation tasks, with an emphasis on the spatial reasoning performance under long-horizon egocentric settings.
    \item We systematically assess the impact of depth-based priors by fusing depth information into RGB inputs, and analyze how such input-level modifications influence the spatial reasoning behavior of VLMs. 
\end{itemize}

%Talk about the Dataset
% Our goal is to determine if VLMs can provide reliable inference for a PVI who enjoys jogging outdoors in urban and off-trail areas. To build our benchmark, we sample 647 real-world Videos from Google's SANPO dataset \cite{waghmare2024sanposceneunderstandingaccessibility}.

% summarize contributions
% Our contributions are as follows:
% \begin{itemize}
%     \item A benchmark for evaluating VLMs assistants for visual navigation safety
%     \item Evaluation of the tradeoffs between small and large parameter VLMs 
%     \item Verification that spatial VLMs have the potential to be superior for navigation guidance. 
% \end{itemize}

% Requires: \usepackage{booktabs}
\begin{table*}[htbp]
    \centering   
    \resizebox{0.8\linewidth}{!}{
    \begin{tabular}{l |cc|cc|cc|cc|cc|cc}
        \toprule
        \textbf{Method}& \multicolumn{2}{c}{\textbf{Rank}}& \multicolumn{2}{c}{\textbf{Avg.}}& \multicolumn{2}{c}{\textbf{Intersection}}& \multicolumn{2}{c}{\textbf{Path}}& \multicolumn{2}{c}{\textbf{Obststructions}}& \multicolumn{2}{c}{\textbf{Pedestrians}}\\ \toprule
 Num. of Choice& \multicolumn{2}{c}{-}& \multicolumn{2}{c}{-}& \multicolumn{2}{c}{2}& \multicolumn{2}{c}{4}& \multicolumn{2}{c}{2}& \multicolumn{2}{c}{2}\\
 With Depth Map&  N&Y&  N&Y&  N&Y&  N&Y&  N&Y& N&Y\\
        \midrule
        VLM-3R & 4 &\textcolor{Green!90}5& 63.52 &\cellcolor{lime!30}64.23& 75.82&\cellcolor{lime!30}75.87& 80.97&\cellcolor{lime!30}81.72& 12.33&\cellcolor{lime!70}15.07& 84.97&\cellcolor{red!10}84.31\\
        VILA-1.5-8B & 2 &\textcolor{red!90}1& 70.10 &\cellcolor{lime!30}70.59& 88.89&\cellcolor{red!10}88.24& 83.96&\cellcolor{red!10}83.21& 21.92&\cellcolor{lime!70}24.66& 85.62&\cellcolor{lime!30}86.27\\
        LLaVA-OneVision-0.5B & 7 &7& 47.84 &\cellcolor{lime!70}48.91& 50.33&\cellcolor{lime!70}56.21& 55.60&\cellcolor{lime!70}58.21& 53.42&\cellcolor{red!50}46.58& 32.03&\cellcolor{lime!70}34.64\\
        LLaVA-OneVision-7B & 6 &6& 52.29 &\cellcolor{red!30}51.13& 58.82&\cellcolor{red!30}57.52& 85.07&\cellcolor{red!10}84.33& 12.33&\cellcolor{lime!30}12.39& 52.94&\cellcolor{red!50}50.33\\
        LLaVA-NeXT-Video-7B & 4 &4& 63.52 &\cellcolor{lime!70}65.64& 75.16&\cellcolor{lime!70}79.08& 83.58&\cellcolor{lime!30}83.63& 12.31&\cellcolor{lime!30}12.42& 83.01&\cellcolor{lime!70}87.58\\
        InternVL2-2B & 1 &\textcolor{Green!90}3& 70.72 &\cellcolor{red!50}67.67& 83.01&\cellcolor{red!10}82.35& 67.16&\cellcolor{red!30}66.42& 43.84&\cellcolor{red!50}35.62& 88.89&\cellcolor{red!50}86.27\\
        InternVL2-8B & 3 &\textcolor{red!90}2& 65.73 &\cellcolor{lime!70}68.09& 81.70&\cellcolor{lime!70}85.62& 79.85&\cellcolor{lime!70}81.34& 16.44&\cellcolor{lime!70}17.81& 84.97&\cellcolor{lime!70}87.58\\
        \bottomrule
    \end{tabular}}
        \caption{Sanpo-D Benchmark}
    \label{tab:vlm_results}
    % \vspace{-0.5cm}
\end{table*}

\begin{figure}[htbp]
    \centering
    \centerline{\includegraphics[width=0.8\linewidth]{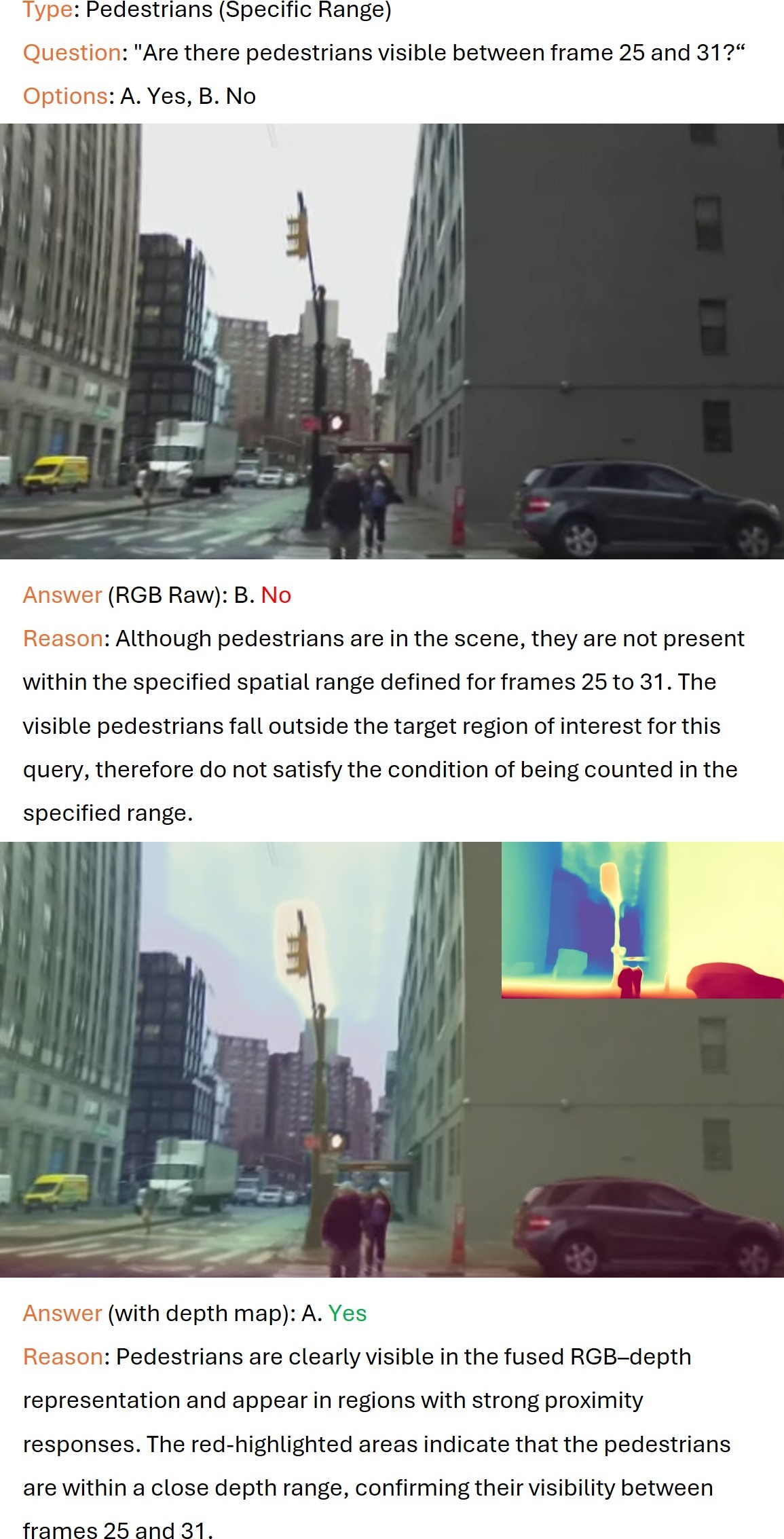}}
    \caption{Example comparison of spatial reasoning with and without depth fusion.}
    \label{fig:example}
\end{figure}

%-------------------------------------------------------------------------
\section{Method}
\label{sec:method}

% Initially, to confirm our hypothesis that spatial VLMs have the potential to outperform traditional ones, we run the navigation-path test from thinking in space\cite{yang2025thinkingspacemultimodallarge} with the keyword "robot" changed to "blind person" on our chosen models. 
% % Explain dataset creation
% For every video, we have a human-annotated Q\&A pair. There are four possible question types. There are 647 Q\&A pairs. 
% I am Jogging, are there any Intersections visible between frame X and Y?
% Are there Pedestrians visible between frames X and Y?
% Are there any obstructions present to my left or right between frame X and Y?
% Where am I currently running between frame X and Y?

Initially, to obtain a preliminary validation of our hypothesis that spatially grounded vision--language models (VLMs) may outperform conventional VLMs on navigation-oriented reasoning, we adopt the navigation-path evaluation protocol proposed in the previous study~\cite{yang2025thinkingspacemultimodallarge}. 
Following the original setup, we apply this test to our selected models with a minimal modification to the prompt, replacing the keyword \textit{'robot'} with \textit{'human'} to better align with visual navigation scenarios, while preserving the original reasoning structure.

For dataset construction, we curate a set of human-annotated question--answer (Q\&A) pairs for each egocentric video. In total, the dataset contains $647$ Q\&A pairs, each associated with a temporal segment defined by a start frame $X$ and an end frame $Y$. The questions are grouped into four navigation-relevant types that require explicit spatial and temporal reasoning:
\begin{itemize}
    \item \textbf{Intersection awareness (153 clips):} ``I am jogging; are there any intersections visible between frame $X$ and $Y$?''
    \item \textbf{Pedestrian presence (268 clips):} ``Are there pedestrians visible between frames $X$ and $Y$?''
    \item \textbf{Obstacle localization (73 clips):} ``Are there any obstructions present to my left or right between frame $X$ and $Y$?''
    \item \textbf{Path context recognition (153 clips):} ``Where am I currently running between frame $X$ and $Y$?''
\end{itemize}

These question types are designed to probe different aspects of spatial reasoning in long-horizon egocentric video, including environmental layout understanding, object presence within a specified temporal window, relative spatial positioning, and scene-level contextual inference.

%-------------------------------------------------------------------------

\section{Results}

\subsection{Sanpo-D Benchmark}
Our preliminary results, summarized in Table~\ref{tab:vlm_results}, reveal a clear trade-off between architectural specialization and general-purpose zero-shot accuracy. High-capacity models such as VILA-1.5-8B and InternVL2-8B achieve the highest overall weighted average accuracies (0.8089 and 0.7817, respectively). In contrast, VLM-3R, despite being an untrained spatial VLM, exhibits a distinct advantage in safety-critical categories. Notably, VLM-3R achieves the highest pedestrian detection accuracy (0.8487) among all evaluated models. This result suggests that the integration of instruction-aligned 3D reconstruction provides a strong inductive bias for spatial reasoning and social awareness, even in the absence of task-specific fine-tuning.

In addition, our results provide empirical support for the hypothesis that smaller models can exhibit stronger generalization capabilities~\cite{barron2025bigthinkcapacitymemorization}. Specifically, LLaVA-OneVision-0.5B and InternVL2-2B demonstrate more consistent performance in obstruction detection compared to larger counterparts such as LLaVA-OneVision-7B. This observation indicates that smaller models may be more effective at extrapolating broad navigational patterns, whereas larger models tend to rely more heavily on memorization of object-specific features.

% This finding is particularly relevant for navigation-oriented and safety-critical applications, as it implies that compact models may generalize more reliably to novel or hazardous environments that are underrepresented or absent in training data. Furthermore, smaller models offer practical advantages in terms of computational efficiency and inference latency, which are critical considerations for real-world deployment.

We further speculate that the superior performance of smaller models on obstruction-related tasks arises from their reliance on abstract spatial regularities rather than explicit object memorization. Larger models may require more direct supervision to associate specific object categories with obstruction semantics, whereas smaller models appear to capture these relationships implicitly through broader spatial patterns~\cite{barron2025bigthinkcapacitymemorization}.
The average accuracy (Avg.) reported in Table~\ref{tab:vlm_results} is computed as the mean accuracy across all evaluated categories.

\begin{figure*}[htbp]
    \centering
    \centerline{\includegraphics[width=1\textwidth]{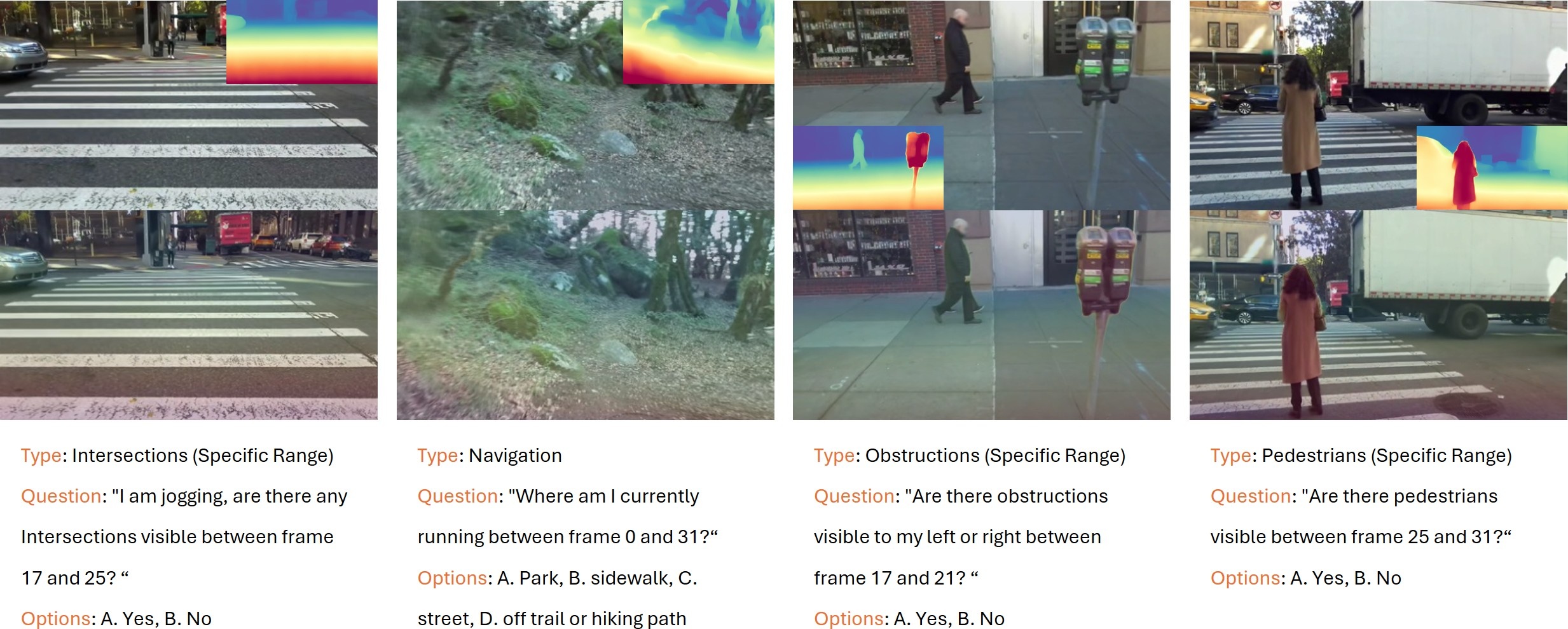}}
    \caption{Sample frames and questions of the proposed Sanpo-D dataset.}
    \label{fig:grid}
    \vspace{-0.25cm}
\end{figure*}

\subsection{Depth-Fusion Experiment}
We further evaluate the effect of explicit depth priors by fusing colorized depth maps with RGB frames and using the resulting representations as inputs to the VLMs, as shown in Figure \ref{fig:grid}. This experiment introduces no changes to model architectures, parameters, or inference procedures. 
As shown in Table~\ref{tab:vlm_results}, depth fusion yields a positive effect for the majority of evaluated models, with particularly noticeable improvements on the obstruction detection task, where many models exhibit relatively weak baseline performance. 
This suggests that explicit depth features can provide complementary spatial information that is difficult to infer from RGB inputs alone. 
However, performance degradation is also observed for certain models, most notably InternVL2-2B and LLaVA-OneVision-7B. 
While the underlying causes of these declines remain unclear, notably, the observed trends do not appear to correlate directly with model parameter scale, indicating that the impact of depth fusion is not solely determined by model size.

While depth information does not uniformly improve accuracy, it consistently alters model behavior, indicating that input-level spatial information introduces a significant inductive bias for spatial reasoning in long-horizon egocentric video, as shown in Figure\ref{fig:example}.

%-------------------------------------------------------------------------
\section{Conclusion}
\label{sec:conclusion}
% In this work, we leveraged the inherent multimodal reasoning of Spatial Vision-Language Models (VLMs) to address the persistent limitations of conventional navigation aids. Our framework leverages the dynamic question-understanding capabilities of these models to help users navigate complex outdoor environments safely. Preliminary results suggest that Spatial VLMs possess the potential to generate personalized, intuitive instructions that align with how humans naturally conceptualize and describe spatial relationships. While our specialized model, VLM-3R, is currently untrained and slightly underperforms in raw accuracy compared to traditional VLMs, such as VILA-1.5-8B and InternVL2-8B, which exhibit exceptional zero-shot performance, it demonstrates significant promise in specific safety metrics, particularly in pedestrian detection, where it achieved an accuracy of 0.8497. 
% We speculate that targeted fine-tuning will provide VLM-3R with a decisive advantage by bridging this initial performance gap while maintaining its specialized spatial focus. Furthermore, our observations indicate that smaller parameter models (e.g., 0.5B) may generalize environmental patterns more effectively than larger counterparts, which often exhibit behaviors indicative of data memorization rather than true extrapolation. While larger models may require exhaustive training on specific obstruction types, smaller models appear more capable of identifying broader navigational patterns, offering a more scalable and computationally efficient path toward real-time assistive technology

In this work, we investigate the role of spatially grounded vision--language models (VLMs) in addressing long-standing limitations of visual navigation systems. By leveraging the multimodal reasoning capabilities of spatial VLMs, we explore their ability to support query-driven understanding of complex outdoor environments without modifying existing inference frameworks. Our preliminary results suggest that spatial VLMs have the potential to generate intuitive and context-aware responses that align with human interpretations of spatial relationships.
% Although the specialized model, VLM-3R, is currently untrained and exhibits slightly lower overall accuracy compared to high-capacity general-purpose models such as VILA-1.5-8B and InternVL2-8B, which demonstrate strong zero-shot performance, it shows clear advantages in specific safety-critical categories. In particular, VLM-3R achieves strong pedestrian detection performance, with an accuracy of 0.8497, indicating that explicit spatial representations can provide meaningful benefits even without task-specific fine-tuning.
% Our empirical observations indicate that smaller parameter models (e.g., 0.5B) may generalize environmental patterns more effectively than larger counterparts, which often exhibit behavior consistent with memorization rather than extrapolation. While larger models may rely on extensive supervision to associate specific object categories with navigational semantics, smaller models appear better suited to capturing broader spatial regularities, offering a more scalable and computationally efficient direction for real-time visual navigation.
Looking forward, while this work focuses on depth fusion as an explicit spatial prior, other complementary information, including motion, optical flow, or geometric consistency across time, may further enhance spatial reasoning when integrated at the input level. Second, long-horizon egocentric video remains inherently challenging due to accumulated viewpoint drift, partial observability, and the absence of persistent global context. A deeper analysis of how these factors affect VLM reasoning over extended temporal spans is needed. Finally, targeted fine-tuning strategies tailored to long-horizon egocentric settings may help models better retain spatial consistency across time, enabling more robust and reliable reasoning in open-ended navigation scenarios.

%-------------------------------------------------------------------------

{
    \small
    \bibliographystyle{ieeenat_fullname}
    \bibliography{main}
}

\end{document}